\documentclass[conference]{IEEEtran}
\IEEEoverridecommandlockouts
\usepackage{cite}
\usepackage{orcidlink}
\usepackage{hyperref}
\usepackage{amsmath,amssymb,amsfonts}
\usepackage{algorithm}
\usepackage{algorithmic}
\usepackage{graphicx}
\usepackage{textcomp}
\usepackage{xcolor}
\usepackage{booktabs}
\def\BibTeX{{\rm B\kern-.05em{\sc i\kern-.025em b}\kern-.08em
    T\kern-.1667em\lower.7ex\hbox{E}\kern-.125emX}}
    
\graphicspath{{figures/}}

\newcommand{\fref}[1]{Fig.~\ref{#1}}
\newcommand{\tref}[1]{Table~\ref{#1}}
\newcommand{\eref}[1]{Equation~(\ref{#1})}
\newcommand{\cref}[1]{Chapter~\ref{#1}}
\newcommand{\sref}[1]{Section~\ref{#1}}

\newcommand{\bs}[1]{\boldsymbol{#1}}
\newcommand\algorithmicprocedure{\textbf{procedure}}
\newcommand{\algorithmicendprocedure}{\algorithmicend\ \algorithmicprocedure}
\makeatletter
\newcommand\PROCEDURE[3][default]{%
  \ALC@it
  \algorithmicprocedure\ \textsc{#2}(#3)%
  \ALC@com{#1}%
  \begin{ALC@prc}%
}
\newcommand\ENDPROCEDURE{%
  \end{ALC@prc}%
  \ifthenelse{\boolean{ALC@noend}}{}{%
    \ALC@it\algorithmicendprocedure
  }%
}
\newenvironment{ALC@prc}{\begin{ALC@g}}{\end{ALC@g}}
\makeatother
    
\begin{document}

\title{A Local Optima Network Analysis of the Feedforward Neural Architecture Space\\
}

 \author{
 \IEEEauthorblockN{Isak Potgieter}
 \IEEEauthorblockA{\textit{Department of Computer Science} \\
 \textit{University of Pretoria}\\
 Pretoria, South Africa \\
 isak.potgieter@gmail.com}
 \and
 \IEEEauthorblockN{Christopher W. Cleghorn}
 \IEEEauthorblockA{\textit{School of Computer Science and Applied Mathematics} \\
 \textit{University of the Witwatersrand}\\
 Johannesburg, South Africa \\
 christopher.cleghorn@wits.ac.za}
 \and
 \IEEEauthorblockN{Anna S. Bosman\orcidlink{0000-0003-3546-1467}}
 \IEEEauthorblockA{\textit{Department of Computer Science} \\
 \textit{University of Pretoria}\\
 Pretoria, South Africa \\
 anna.bosman@up.ac.za}
 }

\maketitle


\begin{abstract}
This study investigates the use of local optima network (LON) analysis, a derivative of the fitness landscape of candidate solutions, to characterise and visualise the neural architecture space. The search space of feedforward neural network architectures with up to three layers, each with up to 10 neurons, is fully enumerated by evaluating trained model performance on a selection of data sets. Extracted LONs, while heterogeneous across data sets, all exhibit simple global structures, with single global funnels in all cases but one. These results yield early indication that LONs may provide a viable paradigm by which to analyse and optimise neural architectures.
\end{abstract}



\section{Introduction}

Much remains unclear about the relationship between neural network architecture, i.e. the number and arrangement of network nodes, and trained model performance. In practise, much of neural architecture optimisation is done empirically. Deeper and, particularly, wider neural networks have been found to be resilient to overfitting \cite{caruana2001overfitting, sagun2014explorations}, improving model capacity to approximate continuous functions \cite{cybenko1989approximation, hornik1989multilayer, lu2017advances} and to consistently converge \cite{bosman2020loss, springer2020s}. Recent findings have shown that larger networks rely on favourably seeded subnetworks \cite{frankle2018lottery, springer2020s}. Lacking \textit{a priori} knowledge about what makes performant subnetworks special, however, such findings may be interpreted as a prescription to train overparameterised networks~\cite{springer2020s}, containing far more parameters than inherently required for optimal training. The gain in model performance comes at a significant computational cost, making deep neural networks time and energy inefficient. This provides motivation to seek new methods with which to characterise the neural architecture space.\looseness=-1

Solutions in a search space and the relationships between them can be characterised by introducing a topology using the notion of a fitness landscape \cite{malan2013survey}, with appropriate definitions of a solution's fitness and its neighbourhood. Ochoa \textit{et al.} \cite{ochoa2008study} introduced a derivative of the fitness landscape, a local optima network (LON), narrowing the focus to local optima, which comprise the LON nodes. Nodes are connected in a directed graph by edges weighted based on the probability of a perturbation from the starting node resulting in an escape to the basin of the destination node \cite{ochoa2008study, mostert2019insights}. LONs have been used to study NK-landscapes \cite{ochoa2008study}, difficulty phase transitions in combinatorial optimisation \cite{ochoa2017understanding}, and to characterise feature selection algorithms \cite{mostert2019insights}. The use of a LON may enhance analysis of the global landscape structure by aiding visual inspection and yielding a range of numerical landscape features. Combined feature assessment of landscape features enhances search space characterisation, search hardness estimation and optimisation. \looseness=-1

As an initial proof of concept of the application of LON analysis to the neural architecture space, the search space considered in this study comprises all possible feedforward neural network architectures with a depth of up to 3 hidden layers, each with up to 10 neurons, trained on a given data set. The fitness of all solutions is measured by taking the mean performance of the corresponding trained model on test data over 30 runs. The evaluation is repeated over 9 different data sets comprising classification and regression tasks, and the results are compared. From each resulting fitness landscape, a LON is derived and its structure investigated to establish the viability of the use of LONs for neural architecture search and optimisation. The following novel results are established:\looseness=-1
\begin{itemize}
	\item A high local optimum fitness standard deviation is related to low modality, which improves the relative ease of fitness optimisation. Low standard deviation at low fitness levels, related to high modality, signals that a model is not sufficiently powerful for a given task.\looseness=-1
	\item Node and edge counts are linearly correlated beyond low extremes. Higher modality involves a higher number of smaller basins, so that a higher proportion of basins are out of reach of any particular candidate solution.
	\item Edges are more likely to be fitness improving than deteriorating, coinciding with a general increase in local optima basin size.\looseness=-1
	\item LONs differ notably between data sets in terms of node and edge count, fitness distribution, basin size distribution and global optimum incoming strength. The solution space is thus heterogeneous and dependent on the particular data set.\looseness=-1
	\item Despite the diverging LON characteristics, all data sets except one produce a single global sink. The simple funnel structure suggests that simple iterative local optimisers may effectively be used to find near optimal solutions.\looseness=-1
\end{itemize}


This study maintains a focus on densely connected feedforward neural networks and does not consider alternative architectural arrangements, such as recurrent or partial connections. The search space is restricted to the neural architecture, excluding the weight space and selection of activation function, while acknowledging that both have been shown to have great bearing on model performance. To control for variance in performance due to randomised weight initialisation, the mean fitness over 30 models trained per architecture is used. The range of architectures evaluated and the data sets used are relatively small in size, due to computational constraints. These selections serve to establish initial viability of local optima network analysis in the present domain by fully enumerating the space over multiple data sets.\looseness=-1

The rest of the paper is organised as follows. Section~\ref{sec:meth:fl} discusses fitness landscapes. Section~\ref{sec:lons:lon} provides the background on LONs. Section~\ref{sec:meth:nn} details the neural network configuration used in the study. Section~\ref{sec:meth:data} discusses the datasets used. Section~\ref{sec:results} presents the empirical results of the study. Finally, Section~\ref{sec:con:res} concludes the paper.\looseness=-1

\section{Fitness Landscapes} \label{sec:meth:fl}


A fitness landscape is formally defined by three components, namely a set of solutions $S$, a notion of a neighbourhood $N(s)$ of each solution $s \in S$, and a fitness function $f(s)$ that assigns a fitness value to each solution \cite{malan2013survey}. The approach of assigning fitness values to candidate solutions has been adapted for optimisation problems. To analyse the interrelation between solutions, the notion of a neighbourhood is introduced based on a suitable distance measure. Thus a topology is constructed, producing a fitness landscape. A landscape is uniquely defined by the tuple $(S, N, f)$. If either the neighbourhood or the fitness function is changed, a different fitness landscape results, even if the set of solutions is maintained. Various landscape features can be extracted, such as the fitness distribution, the funnel structure, and the number and structure of optima.\looseness=-1

Intuitively, a fitness landscape corresponds to a physical landscape, where the relative `heights' of fitness values shape hills, plateaus, sinks and valleys. In the context of search algorithms, the fitness landscape can be formalised as a graph with solutions as its nodes and edges between neighbours \cite{malan2013survey}. The neighbourhood of a solution $s$ is the set of solutions obtained by applying a search operator, starting with $s$ \cite{moser2017identifying}. In discrete space, as is the case with neural architectures, the operator may be defined as a reconfiguration of solution parameters, such as a bit flip in a binary string \cite{ochoa2008study, ochoa2017understanding}. In continuous space, the operator may be described using a distance metric, such as Euclidean distance \cite{malan2013survey}. \looseness=-1

In order to consider neural architecture space from the fitness landscape analysis perspective, definitions of candidate solutions, their fitness and neighbourhoods, as well as what constitutes a local minimum, must be provided. These are given in the subsections below.\looseness=-1

\subsection{Candidate solutions} \label{sec:meth:fl:sol}

The solution space $S$ considered in this study comprises all possible feedforward neural network architectures with a depth of up to 3 hidden layers, each comprising between 1 and 10 neurons, trained on a given data set. For ease of reference, neural networks with $n$ \textit{hidden layers} are here simply referred to as having $n$ \textit{layers}, while the full network invariably contains input and output layers. Only the network architecture is evaluated, randomly initialising weights and keeping other parameters, such as activation functions and their configurations, fixed. The neural architecture can be depicted as the set of 1-, 2- and 3-tuples of numbers 1 to 10. More formally, given $A = \{1, 2, \dots, 10\}$, the solution space is parametrised as $S' = (A)\; \cup\; (A \times A)\; \cup\; (A \times A \times A)$ \footnote{where $\times$ denotes the cartesian product and $\cup$ set union}. For example, $(3) \in (A)$, $(4, 6) \in (A \times A)$ and $(5, 8, 2) \in (A \times A \times A)$.\looseness=-1

The architecture search space is combinatorial, being discrete and finite. Within a defined maximum layer depth $d$ and individual layer width $w$, $|S| = \sum_{l=1}^{d}w^l$. With $d=3, w=10$ the total amounts to $|S| = |S'| = 1110$. $S$ is the set of feedforward networks parametrised as $S'$, which can be expressed as $S = \mathcal{X}(S')$, where $\mathcal{X}$ represents the derivation of the architecture instantiations corresponding to parametrised solutions in $S'$. Denote positional digits of $n$-tuple $s' \in S'$ by $(s'_1,\dots, s'_d)$, a hidden layer containing $m$ neurons by $\bs{h}[m]$ and a directional connection between layers by $\bs{h}\rightarrow\bs{h}$. Accordingly, a solution $s \in S$ can be constructed as
\begin{equation*}
\begin{split}
	s = \mathcal{X}(s') & = \mathcal{X}((s'_1, \dots, s'_d)) \\
	& = \bs{x} \rightarrow \bs{h}[s'_1] \rightarrow \dots \rightarrow \bs{h}[s'_d] \rightarrow \bs{y}.
\end{split}
\end{equation*}

\begin{figure}
	\centering{\includegraphics[width=0.5\columnwidth]{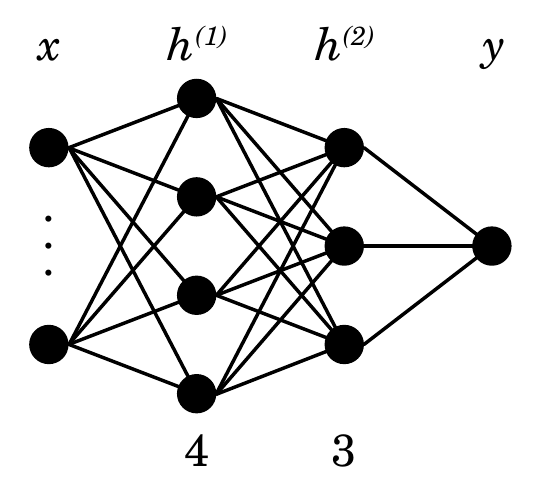}}
	\caption{Model with layers containing 4 and 3 hidden neurons}
	\label{fig:ffnn43}
\end{figure}

For example, if $s' = (4, 3)$, then $s = \mathcal{X}((4, 3)) = \bs{x} \rightarrow \bs{h}[4] \rightarrow \bs{h}[3] \rightarrow \bs{y}$, as depicted in \fref{fig:ffnn43}. The architecture space is fully enumerated by fitting all possible configurations within the defined bounds, training the corresponding model and evaluating its test performance.

\subsection{Fitness evaluation} \label{sec:meth:fl:fit}

As the fitness value of a candidate solution, $f(s)$, we take the performance per configuration, computed as the R-squared ($R^2$) statistic for both regression and classification tasks, defined as
\begin{equation}
	R^2 = 1 - \frac{\sum_i(y^{(i)} - \hat{y}^{(i)})^2}{\sum_i(y^{(i)} - \bar{y})^2},
\end{equation}
computed over $N$ predictions, where $y^{(i)}$ are the actual values of the response variable, $\bar{y}$ is the mean of the response values, and $\hat{y}^{(i)}$ are the corresponding prediction values. $R^2$ values typically range from a maximum of 1 for perfect predictions to 0, but exceptionally large errors can yield arbitrarily large negative values. $R^2$ is selected due to its scale invariance, which aids interpretation of results. To facilitate the use of $R^2$ in classification tasks, the uniform average for multiple outputs (target classes) is used per prediction to obtain residuals. A consistent fitness measure is, therefore, used throughout the study.

For each architecture, a batch of 30 models are trained and evaluated to control for variability in model performance due to random weight initialisation. The first model per batch is identically seeded, and subsequent training procedures correspondingly inherit from it. Thus the first 30 models trained under the same conditions will always produce the same set of results, while individual runs within a batch produce varying results. Controlled seeding is necessary for commensurability between solutions. The mean $R^2$ value per batch is used as the final fitness value per solution.


\subsection{Neighbourhoods} \label{sec:meth:fl:nbh}

Let the neighbourhood $\mathcal{N}(s)$ of a solution $s \in S$ be defined by an operation which either offsets the width of one layer by one or alternatively prunes or clones a layer, within allowed bounds. That is, the neighbourhood of solution $s$ includes all arrangements with either the number of neurons incremented or decremented in one layer, or one layer duplicated or removed, without adjusting layer-wise neuron counts.

\subsubsection{Width Offsets}

Width offsets increment or decrement one layer's neuron count. Let $\mathcal{W}(s)$ denote the set of all possible layer-wise neural decrements and increments applied to a solution $s$. With $\mathcal{W'}(s')$ as the set of offsets of the corresponding $n$-tuple $s' \in S'$, let $N = \{1, \dots, n\}$, width offsets are formalised by $\mathcal{W}(s) = \mathcal{W}(\mathcal{X}(s)) = \mathcal{X}(\mathcal{W'}(s'))$, and
\begin{equation}
\label{eq:woff}
\begin{split}
	\mathcal{W'}(s') & = D \cup I, \text{ where} \\
	D & = \{ t' \in S' \mid \exists ! i \in N \text{ s.t. } t_i' = s'_i-1, t_{j \neq i}' = s'_j  \}, \\
	I & = \{ t' \in S' \mid \exists ! i \in N \text{ s.t. } t_i' = s'_i+1, t_{j \neq i}' = s'_j  \}.
\end{split}
\end{equation}
where $t' \in S'$ implies that width bounds are not exceeded and the corresponding architecture must be a valid solution in $S$, and $\exists!i$ denotes \textit{`there exists a unique $i$'}.

\subsubsection{Depth Offsets}

Depth is offset via either cloning or pruning. Cloning involves replicating the neuron count in an existing layer and adjacently inserting an additional layer. Pruning is done by removing a selected layer, provided at least one layer remains. Due to pruning, neighbourhoods are not necessarily symmetric, e.g. $(2, 2)$ is a neighbour of $(2, 2, 8)$ by virtue of pruning the third layer, but no reverse permutation is permitted. Let $\mathcal{D}(s)$ denote the set of all single layer pruning and cloning permutations to a solution $s$, and $\mathcal{D'}(s')$ the corresponding set resulting from adding to or removing an entry from the $n$-tuple $s'$. The depth offset operation is formalised as $\mathcal{D}(s) = \mathcal{D}(\mathcal{X}(s)) = \mathcal{X}(\mathcal{D'}(s'))$, where
\begin{equation}
\label{eq:doff}
\begin{split}
	\mathcal{D'}(s') & = \{t' \in S' \mid \exists ! i \in N \text{ s.t. } t' = s'_{1,\dots,i,i,\dots,n} \} \\
	& \cup \{t' \in S' \mid \exists ! i \in N \text{ s.t. } t' = s'_{1,\dots,i-1,i+1,\dots,n} \}.
\end{split}
\end{equation}

The resulting set of $n$-tuples in $S'$ must correspond to valid solutions in $S$. Using \eref{eq:woff} and \eref{eq:doff}, the neighbourhood of a solution $s$ is defined as
\begin{equation}
\label{eq:nbh}
\begin{split}
	\mathcal{N}(s) & = \mathcal{W}(s) \cup \mathcal{D}(s) \\
	& = \mathcal{X}(\mathcal{W'}(s')) \cup \mathcal{X}(\mathcal{D'}(s')).
\end{split}
\end{equation}

\begin{figure*}
	\makebox[\textwidth][c]{\includegraphics[width=0.95\textwidth]{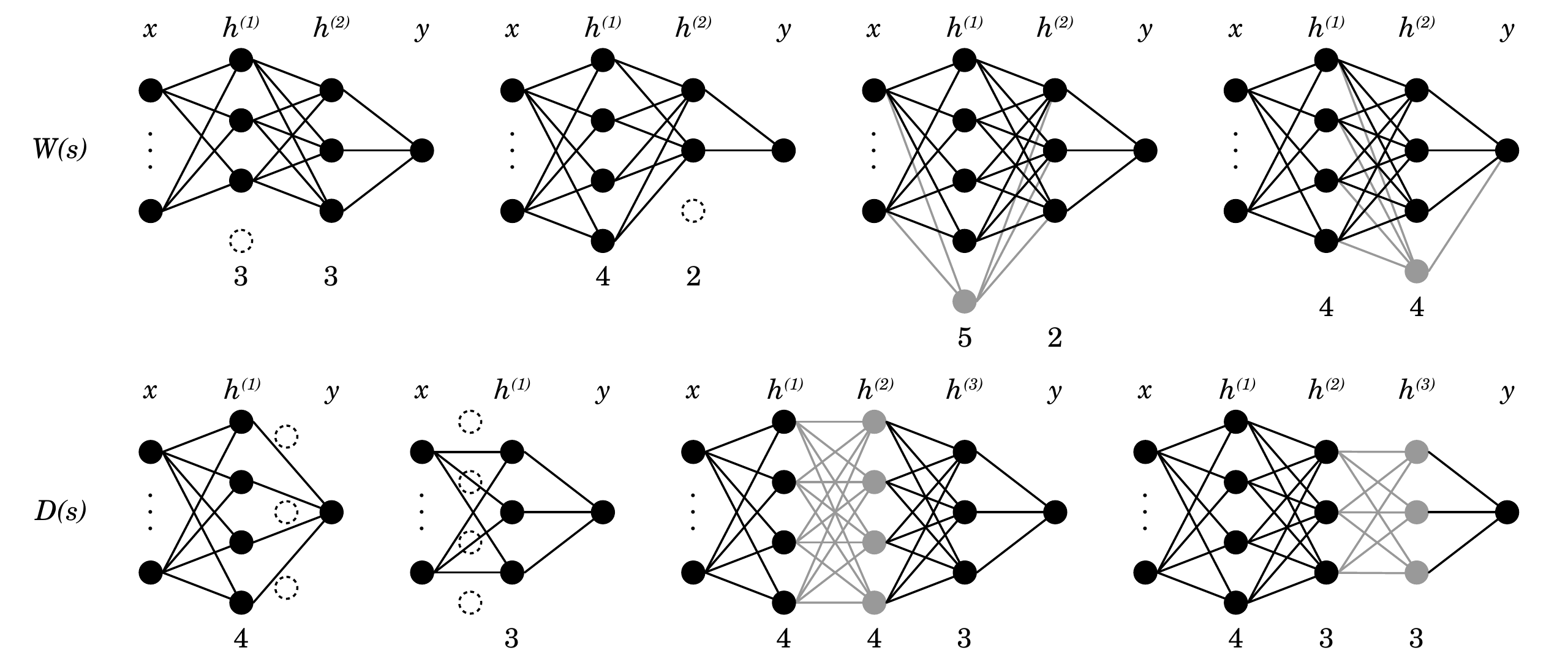}}
	\caption[Neighbourhood of solution $s$ with (4, 3) architecture]{$\mathcal{N}(s)$, where $s' = (4, 3)$}
	\label{fig:ns}
\end{figure*}

\fref{fig:ns} shows an example of the neighbourhood of the solution $s' = (4, 3)$ (illustrated in \fref{fig:ffnn43}), differentiating $W(s)$ and $D(s)$. The neighbourhood operation produces 5879 edges per landscape, and the average number of neighbours per solution\footnote{every edge adds a neighbour to both solutions or nodes it connects, thus twice the edge count is used when computing the average: $5879(2) / 1110$} is around 10.6.

\subsection{Local Optima} \label{sec:lons:lop}

A local optimum (LO) is a solution with a fitness value that matches or exceeds that of all solutions in its neighbourhood. Each local optimum $i$ is surrounded by a basin of attraction, which is the set of solutions from which $i$ will be reached by following a local search procedure. Formally, $b_i = \{ s \in S | h(s) = i \}$, and $|b_i|$ is the cardinality of the basin. A best-improvement hill climbing local search is used in this study to locate local optima, defined in Algorithm~\ref{alg:best}.


\begin{algorithm}
   \caption{Maximising best-improvement hill climbing}
   \label{alg:best}
\begin{algorithmic}
   \STATE {\bfseries Input:} initial solution $x$
   \REPEAT
   \STATE $x' \leftarrow argmax_{y\in\mathcal{N}(x)}f(y)$
   \IF{$f(x') > f(x)$}
   \STATE $x \leftarrow x'$
   \ENDIF
   \UNTIL{$x$ is a local optimum}
\end{algorithmic}
\end{algorithm}

\section{Local Optima Networks} \label{sec:lons:lon}
This study aimed to construct and investigate LONs for the neural architecture space.
For each data set considered, the architecture space was fully enumerated, and the hill climbing terminus starting with each solution was determined, exhaustively populating LON \textit{nodes}. To draw \textit{edges}, the neighbourhood operation defined in \eref{eq:nbh} was selected as a basis, applied by perturbation strength $D = 2$, which is consistent with related literature \cite{ochoa2017understanding, mostert2019insights}, and is also empirically found to be sufficient for escape, particularly from suboptimal solutions. An edge thus represents the relative likelihood of moving into a neighbouring basin of attraction from a local optimum after a controlled perturbation of up to two random moves. The value of $D$ has a marked effect on the connectedness of the LON graph, since more random moves allow for farther reaching escape routes. Directed LON edges are weighted by the number of distinct paths resulting in a move between basins of the two local optima it connects. When calculating edge weights the edge counts between nodes are not scaled to a proportion of all distinct edges from the source node. Thus weights do not represent probabilities, but rather the relative likelihood of escape.

\subsection{Funnels} \label{sec:lons:funn}

In a monotonic LON (MLON) \cite{ochoa2017understanding}, non-deterioration is required for an edge to be drawn between nodes. MLONs yield the notion of funnels, which are the set of monotonic sequences leading to a particular local optimum. A funnel is a basin at the local optima level \cite{thomson2020inferring}, with a sink node as its terminus. MLON sinks have out degree zero, and sources have in degree zero.


\section{Neural Network Configuration} \label{sec:meth:nn}

In the present context, the selection of hyper-parameter values as such is not as important as their consistency, to ensure commensurability of solutions between data sets. Apart from output activation and loss, all parameters were consistent between classification and regression tasks. All configuration settings used in the experiments  are presented in \tref{tbl:nn}.

\begin{table}[t]
\caption{Neural Network Parameters}
\label{tbl:nn}
\vskip 0.15in
\begin{center}
\begin{small}
\begin{sc}
\begin{tabular}{lll}
\toprule
\textbf{Parameter} & \textbf{Classification} & \textbf{Regression} \\
\midrule
Weight init. & \textit{He uniform var.} & \textit{He uniform var.} \\
Hid. layer act. & \textit{ReLU} & \textit{ReLU} \\
Output act. & \textit{Softmax} & \textit{Linear} \\
Loss & \textit{S.c. cross-entropy} & \textit{MSE} \\
Max epochs & 100 & 100 \\
Early stop $\Delta$ & $<0.0001 \times 10$ & $<0.0001 \times 10$ \\
Optimiser & \textit{Adam} (0.01 l.r.) & \textit{Adam} (0.01 l.r.) \\
Train-val-test & 0.7 / 0.15 / 0.15 & 0.7 / 0.15 / 0.15 \\
\bottomrule
\end{tabular}
\end{sc}
\end{small}
\end{center}
\vskip -0.1in
\end{table}


\section{Data} \label{sec:meth:data}

Models were trained on 9 different data sets: 6 classification tasks (\tref{tbl:cls}) and 3 regression tasks (\ref{tbl:reg}). While presenting a small sample, the intention was to analyse the fitness landscapes produced for a range of relatively common data sets. Classification sets included commonly used tabular and image recognition data. While these are relatively small data sets, the two categories differ significantly in kind and difficulty. All three regression sets are tabular, but have diverging input features and pattern counts.

With a limited model and data set size, each model was trained using a single CPU. Over an average of 50.86 epochs, the time required per model was an average of 30.77 seconds,  ranging from 1.71 second for Iris to over 112 seconds for MNIST. Full enumeration of the solution space is, however, non-negligible. For each data set, each of the 1110 architectures was trained 30 times, totalling 33300 training procedures per set. To train the full compliment, a computing cluster\footnote{\url{https://www.chpc.ac.za/index.php/resources/lengau-cluster}} was used to maximise the number of simultaneous training procedures. This consideration limits the number and size of the sets that can feasibly be included.

\begin{table}[t]
\caption{Classification Data Sets}
\label{tbl:cls}
\vskip 0.15in
\begin{center}
\begin{small}
\begin{sc}
\begin{tabular}{llccc}
\toprule
\textbf{Name} & \textbf{Type} & \textbf{Patterns} & \textbf{Inputs} & \textbf{Cls} \\
\midrule
Iris\cite{fisher1936use} & Tab & 150 & 4 & 3 \\
Wine\cite{forina1988parvus} & Tab & 178 & 13 & 3 \\
MNIST\cite{lecun1998gradient} & Img & 70000 & $28\times28$ & 10 \\
F. MNIST\cite{xiao2017fashion} & Img & 70000 & $28\times28$ & 10 \\
CIFAR-10\cite{krizhevsky2017imagenet} & Img & 60000 & $32\times32$ & 10 \\
CIFAR-100\cite{krizhevsky2017imagenet} & Img & 60000 & $32\times32$ & 100 \\
\bottomrule
\end{tabular}
\end{sc}
\end{small}
\end{center}
\vskip -0.1in
\end{table}

\begin{table}[t]
\caption{Regression Data Sets}
\label{tbl:reg}
\vskip 0.15in
\begin{center}
\begin{small}
\begin{sc}
\begin{tabular}{llcc}
\toprule
\textbf{Name} & \textbf{Type} & \textbf{Patterns} & \textbf{Inputs} \\
\midrule
Wine Quality\cite{cortez2009modeling} & Tabular & 4898 & 11 \\
Insurance\cite{choi2017med} & Tabular & 1338 & 6 \\
Housing\cite{belsley2005regression} & Tabular & 506 & 13 \\
\bottomrule
\end{tabular}
\end{sc}
\end{small}
\end{center}
\vskip -0.1in
\end{table}

Image data arrays were flattened by concatenating the rows into a single input vector. Colour values were first converted to greyscale before being flattened. Numeric input variables were $z$-score normalised or standardised. Binary input variables were encoded as $-1$ and $1$, and multiple class inputs were one-hot encoded. Regression labels were left unscaled.


\section{Results}\label{sec:results}

Salient results relating to the extracted fitness landscapes  are presented in \sref{sec:res:fl}, and local optima networks in \sref{sec:res:lon}. Data set names are abbreviated.

\subsection{Fitness Landscapes} \label{sec:res:fl}

\begin{figure}[ht]
\vskip 0.2in
\begin{center}
\centerline{\includegraphics[width=1.02\columnwidth]{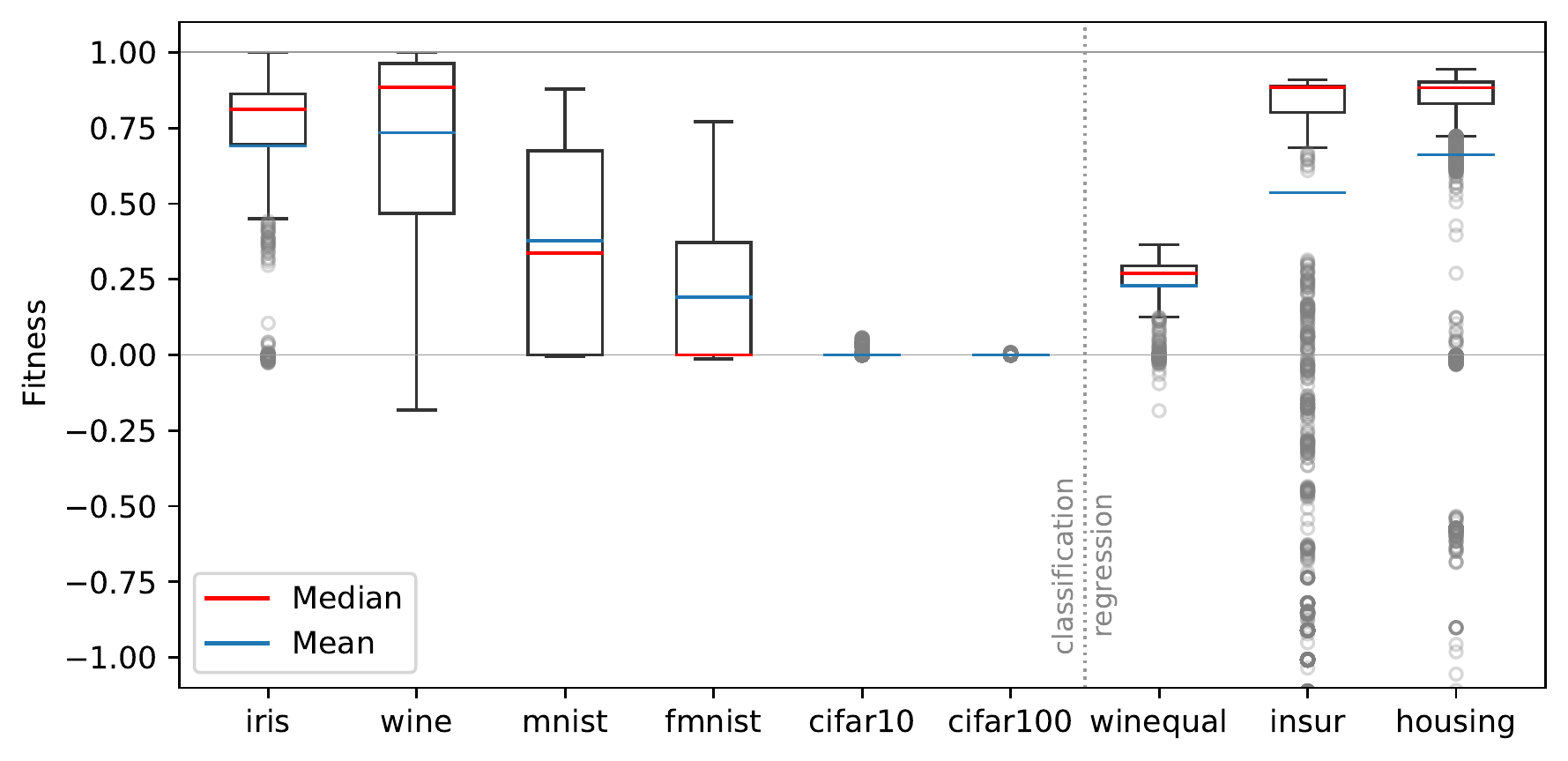}}
	\caption[Solution fitness distribution]{Solution fitness distribution between the evaluated data sets. Boxes cover the interquartile range (IQR), whiskers up to 1.5 times the IQR and circles depict outliers beyond that point. Medians are coloured red and means, blue.}
\label{fig:fitdisp}
\end{center}
\vskip -0.2in
\end{figure}

A bird's eye view of the fitness values produced by the trained neural networks, as represented by the box plots in \fref{fig:fitdisp}, reveals a wide range of outcomes across the different data sets. Only \textit{iris} and \textit{wine} have solutions that achieved an $R^2$ score of 1. Fitness values varied significantly between data sets in terms of mean and median statistics, range and distribution.

For all data sets, the bulk of the fitness values are bounded by 0 and 1, but all sets exhibited solutions with negative fitness, representing a complete failure to fit the response data. Negative outliers were much more extreme for the regression tasks, in particular \textit{insur} and \textit{housing}. \textit{Softmax}, used for classification output, produced a probability distribution of values between 0 and 1 for all classes, summed to a total value of 1, effectively bounding the numeric divergence possible when the target is numeric and of arbitrary scale, as in this study.


\subsection{Local Optima Networks} \label{sec:res:lon}

A LON was extracted for each data set, with local optima (LO) found using best-improvement hill climbing as the nodes and edges drawn between LO if any controlled perturbation to one local optimum (source) reached a solution within the basin of another local optimum (target), as described in \sref{sec:lons:lon}. \tref{tbl:lonfeat} provides a comparison of basic features describing the resulting networks, further illustrating the variation between fitness landscapes.

\begin{table}[t]
\caption{LON features, using abbreviations \textbf{GO} = Global Optima, \textbf{LO} = Local Optima, \textbf{Edg} = Edges, \textbf{Fnl} = Funnels. Task types are indicated by \textit{cls} = classification and \textit{reg} = regression. Global optimum values are rounded to 5 decimals.}
\label{tbl:lonfeat}
\vskip 0.15in
\begin{center}
\begin{small}
\begin{sc}
\begin{tabular}{lclccc}
\toprule
\textbf{Data Set} & \textbf{Task} & \textbf{GO} & \textbf{LO} & \textbf{Edg} & \textbf{Fnl} \\
\midrule
iris     & cls & 1 (0.86255)  & 35      & 619   & 1 \\
wine     & cls & 1 (0.96343)  & 25      & 287    & 1 \\
mnist    & cls & 1 (0.8544)     & 4      & 12     & 1 \\
fmnist   & cls & 1 (0.60385)   & 11      & 56    & 1 \\
cifar10  & cls & 1 (0.00552) & 90    & 2542    & 2 \\
cifar100 & cls & 1 (0.00055) & 101  & 3134  & 1 \\
winequal & reg & 1 (0.30812)  & 26     & 341   & 1 \\
insur    & reg & 1 (0.88873)  & 45      & 974   & 1 \\
housing  & reg & 1 (0.90936)  & 28     & 366    & 1 \\
\bottomrule
\end{tabular}
\end{sc}
\end{small}
\end{center}
\vskip -0.1in
\end{table}

\subsubsection{Local Optima Fitness Distribution}

\begin{figure}[t]
\vskip 0.2in
\begin{center}
\centerline{\includegraphics[width=\columnwidth]{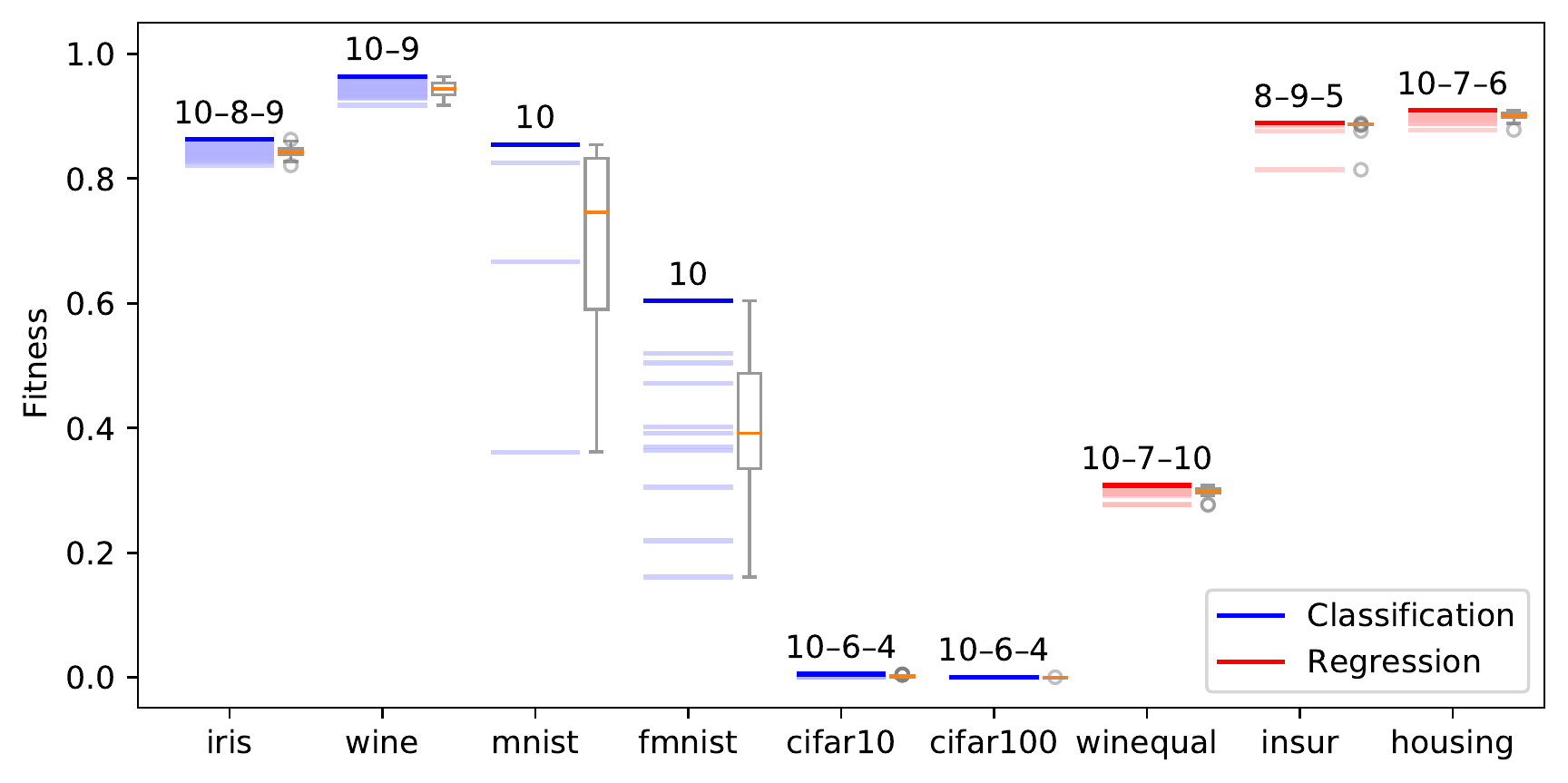}}
	\caption{Local optima fitness comparison between data sets. The globally optimum architecture is indicated above each fitness stack.}
	\label{fig:lops}
\end{center}
\vskip -0.2in
\end{figure}

A comparison of LO fitnesses is displayed in \fref{fig:lops}, demonstrating an array of model capacities relative to the provided data sets. The distributions of LO fitness values also bares resemblance to the distributions of the solution space as a whole (\fref{fig:fitdisp}). The choice of $R^2$ as the fitness metric resulted in all solution fitnesses being globally unique. Consequentially, all data sets produced a unique global optimum (\tref{tbl:lonfeat}).

\subsubsection{Network Modality} \label{sec:res:nodeedge}

\begin{figure}[t]
\vskip 0.2in
\begin{center}
\centerline{\includegraphics[width=0.9\columnwidth]{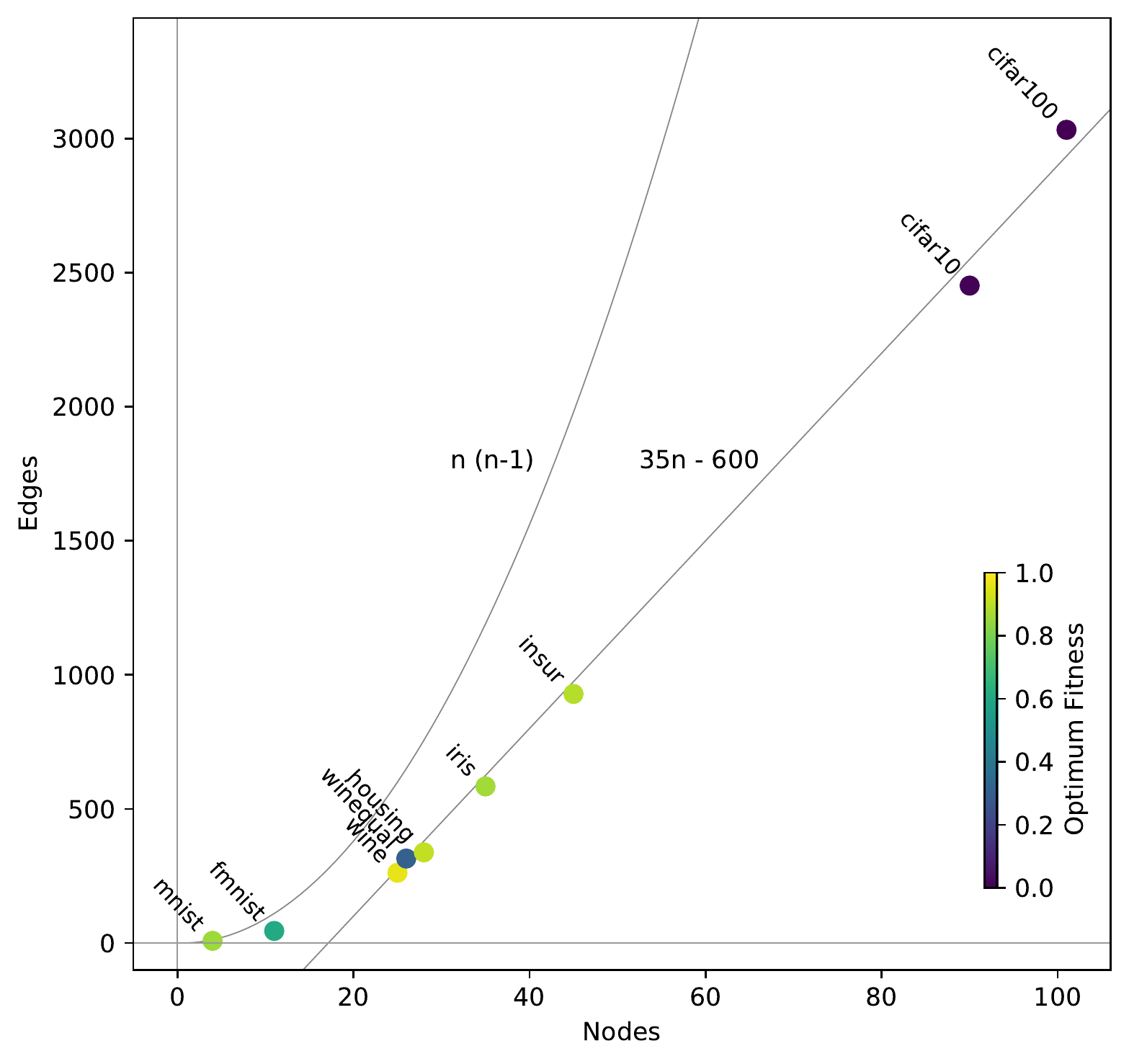}}
	\caption{Relation between LON nodes and edges, coloured according to global optimum value. Additional plots of the maximum possible number of edges, $n(n-1)$, and the actual approximation, $35n-600$, where n is the node count.}
	\label{fig:nodeedgeA}
\end{center}
\vskip -0.2in
\end{figure}

The number of local optima varied from as few as 4 for \textit{mnist} to as many as 101 for \textit{cifar100} -- more than $9\%$ of the search space. As can be expected, the edge count correlated with the number of nodes that they connect, as demonstrated in \fref{fig:nodeedgeA}. While the maximum number of potential directional connections between nodes increased exponentially as the number of nodes increased, totalling $n(n-1)$ for $n$ nodes, it was observed that for the selected data sets, the edge count increased in a linear fashion for networks with more than 20 nodes, approximated by $35n-600$. A possible explanation is that higher modality implies a higher number of smaller basins, so that on average a higher proportion of basins are out of reach from any particular solution.

\begin{figure}[t]
\begin{center}
\centerline{\includegraphics[width=0.9\columnwidth]{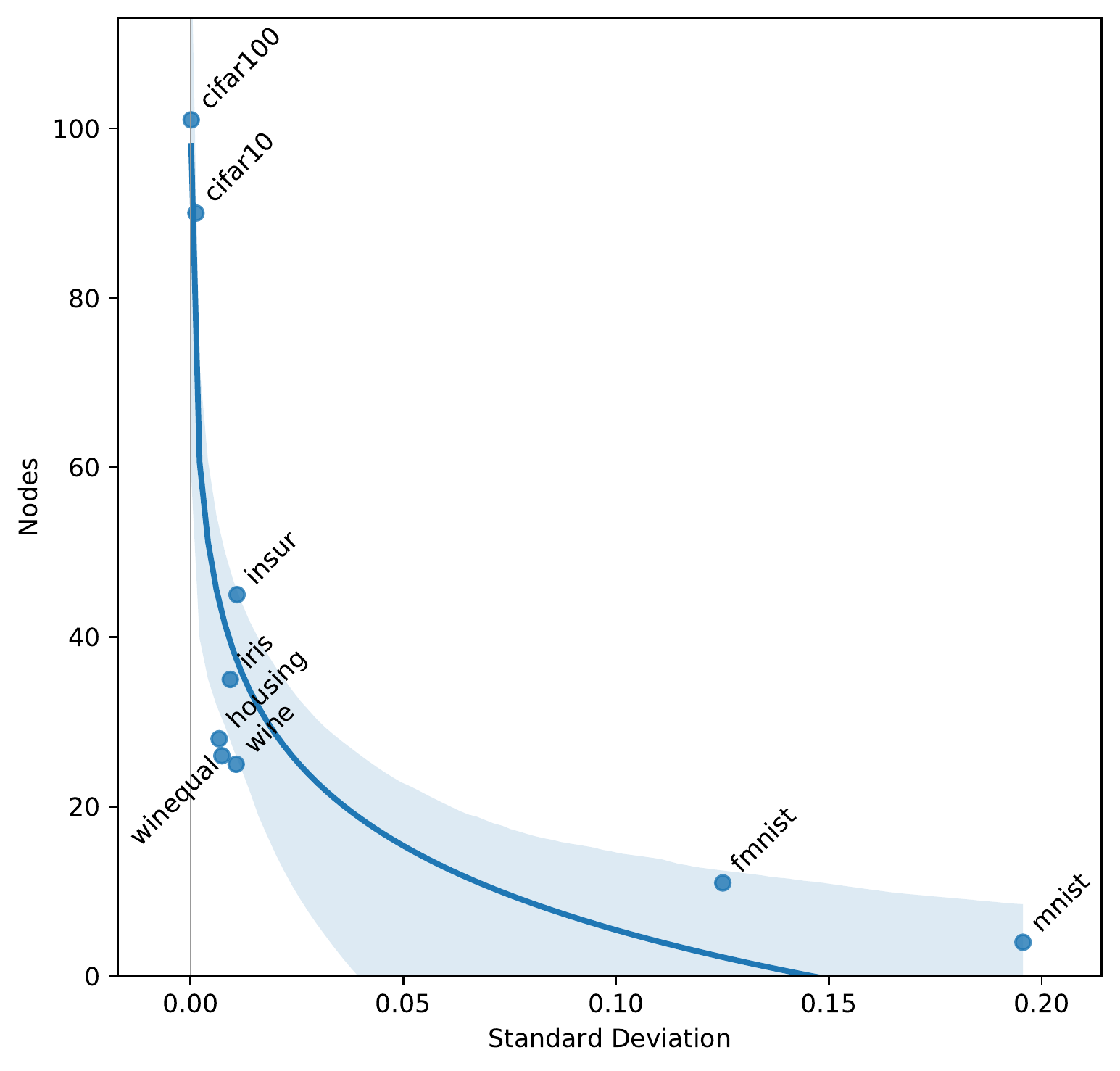}}
	\caption{Relation between fitness SD and node count}
	\label{fig:sdnodeedge}
\end{center}
\vskip -0.2in
\end{figure}

A lower global optimum value reduced the fitness range. This is correlated with an increase in modality when comparing the \textit{cifar} and \textit{mnist} data sets, but is not consistent, as illustrated in \fref{fig:nodeedgeA}. A better indicator may be the fitness standard deviation (SD) between local optima. \fref{fig:sdnodeedge} depicts the number of network nodes over the fitness SD per data set. At the extremes, \textit{cifar100} had the lowest SD and the highest modality, while \textit{mnist} expressed the inverse. The allocation of the intermediate cases made the overall trajectory resemble exponential decay, with the node and edge count decreasing rapidly with increases near the lower end of the SD range. Yet there was a wide spread, for example, between \textit{insur} and \textit{wine}, with similar SD. For the data sets considered, the inverse relation between SD and node count was thus found to be clearest at the extremes.

Since the LON edges are directed, the breakdown of improving and deteriorating connections can be compared, that is, whether following an edge results in a move to a local optimum with a better or worse fitness. On average, improving moves were found to be around 1.8 times more common.

\subsubsection{Local Optima Basins}

\begin{figure}[t]
\vskip 0.1in
\begin{center}
\centerline{\includegraphics[width=\columnwidth]{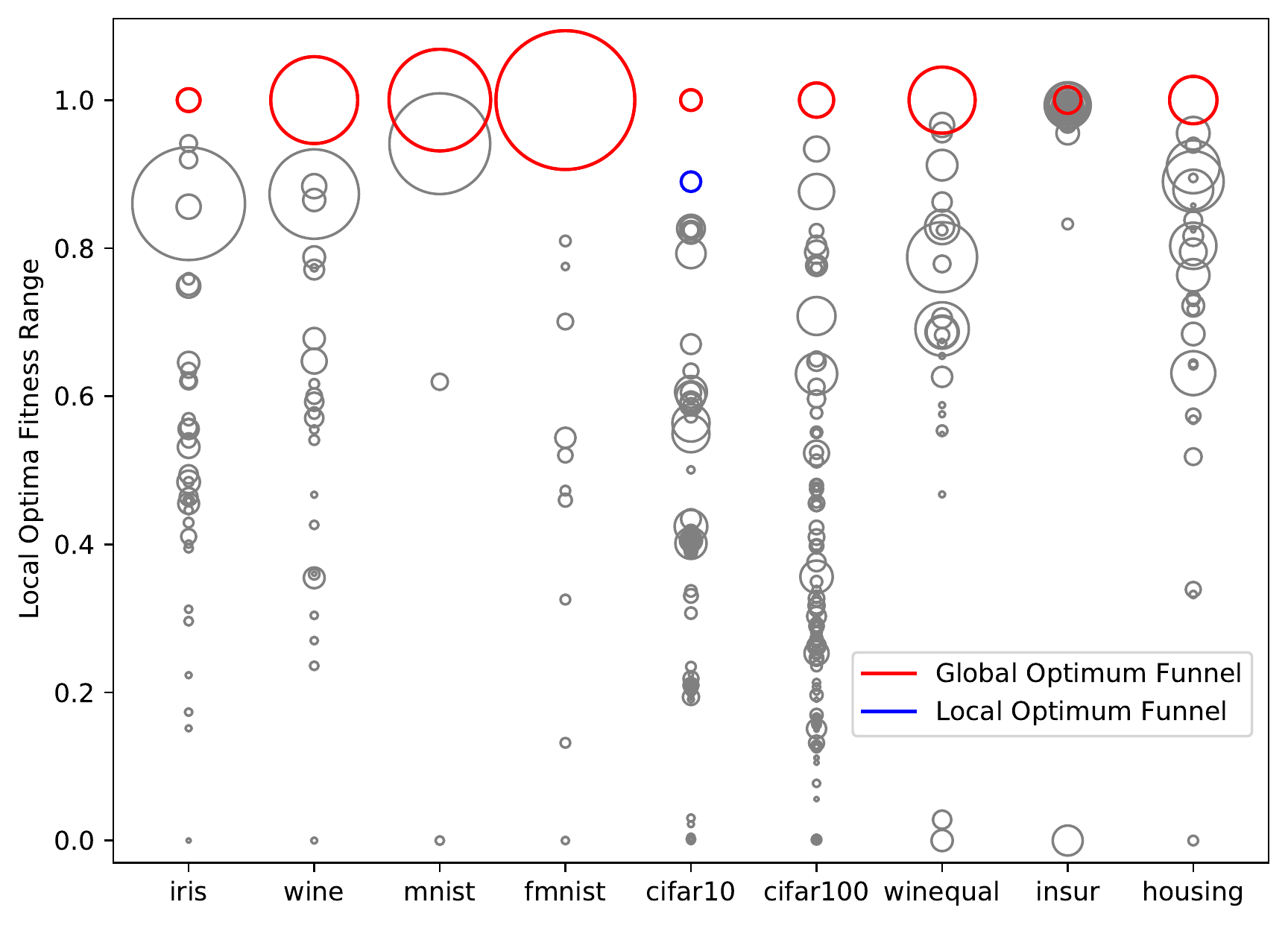}}
	\caption{Local optima basin sizes across relative fitness levels}
	\label{fig:lopfitbas}
\end{center}
\vskip -0.2in
\end{figure}

A predominantly improving edge set makes sense when considering the effect of relative basin sizes between solutions. Fitter solutions can generally be expected to have bigger basins, easier to reach, but harder to escape from. \fref{fig:lopfitbas} provides an overview of the relative LO basin sizes across fitness values per data set. Overall, relative basin size can be observed to scale up as fitness increases. The most extreme example is \textit{fmnist}, which has an absolutely dominant basin at its global optimum.

\begin{figure}[t]
\vskip 0.1in
\begin{center}
\centerline{\includegraphics[width=0.9\columnwidth]{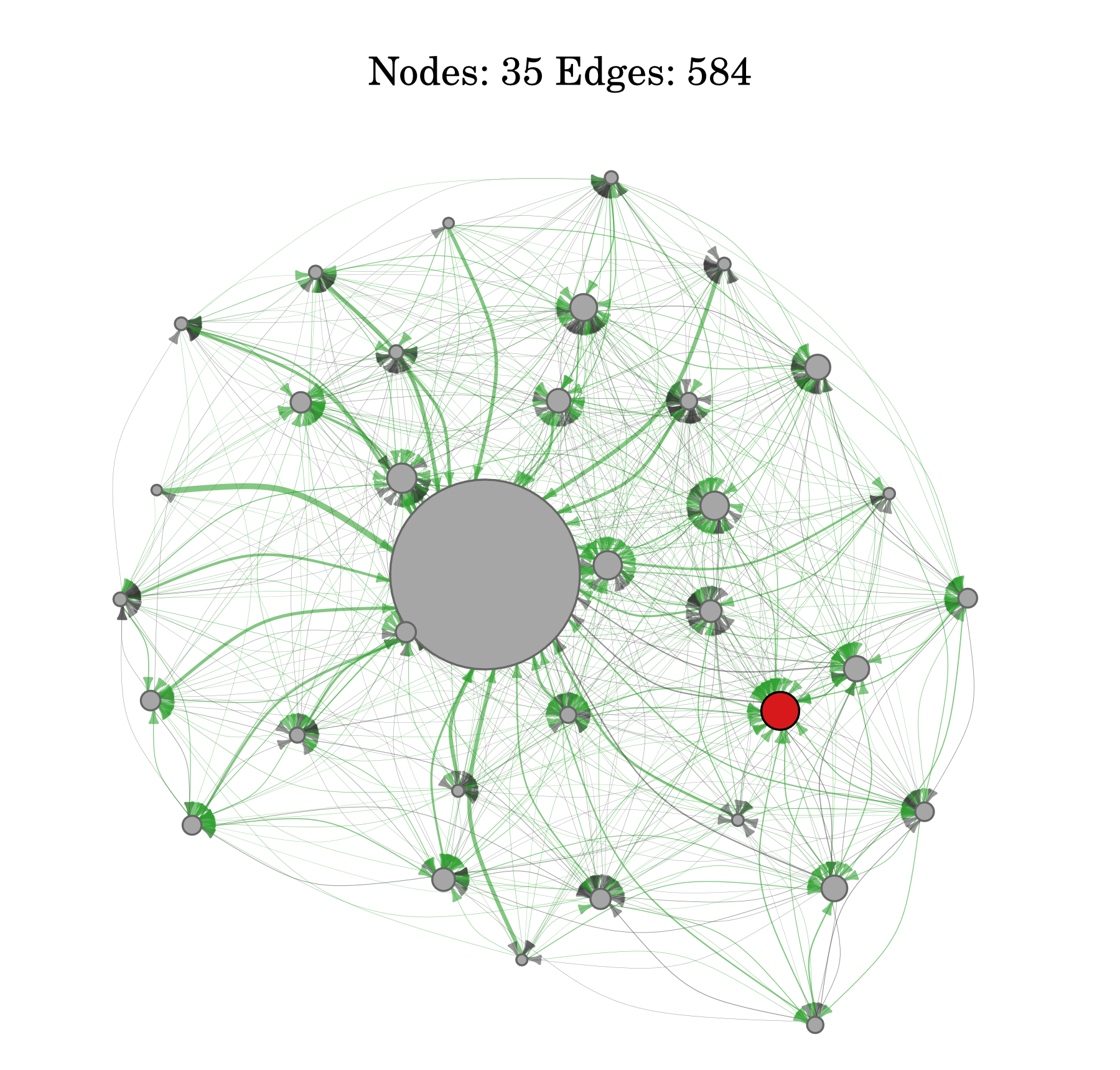}}
	\caption{2-dimensional LON plot for \textit{iris} with nodes scaled by incoming strength or the sum of the incoming edges to a node. Improving edges are coloured green and deteriorating edges are dark grey, with thickness corresponding to edge weight. The global sink is coloured red and all other local optima are grey.}
	\label{fig:irisgraphA}
\end{center}
\vskip -0.2in
\end{figure}

Sizeable suboptimal basins also exist, sometimes exceeding the basin size at the global optimum. \fref{fig:irisgraphA} plots the 2-dimensional LON graph for \textit{iris}, with self loops removed to simplify the layout. The size of each vertex is determined by the incoming strength of LON edges, demonstrating the correspondence between basin size and likelihood of being reached by a perturbation. For the same reason that larger basins at higher fitness levels make improving moves more likely, the basins of suboptimal solutions may provide obstacles to optimisation over the fitness landscape, even forming suboptimal sinks. Despite the prominence of this solution in the LON, it is not a sink, and is outward connected to the comparatively modest global optimum.

\subsubsection{Funnels}

\begin{figure}[t]
\begin{center}
\centerline{\includegraphics[width=0.9\columnwidth]{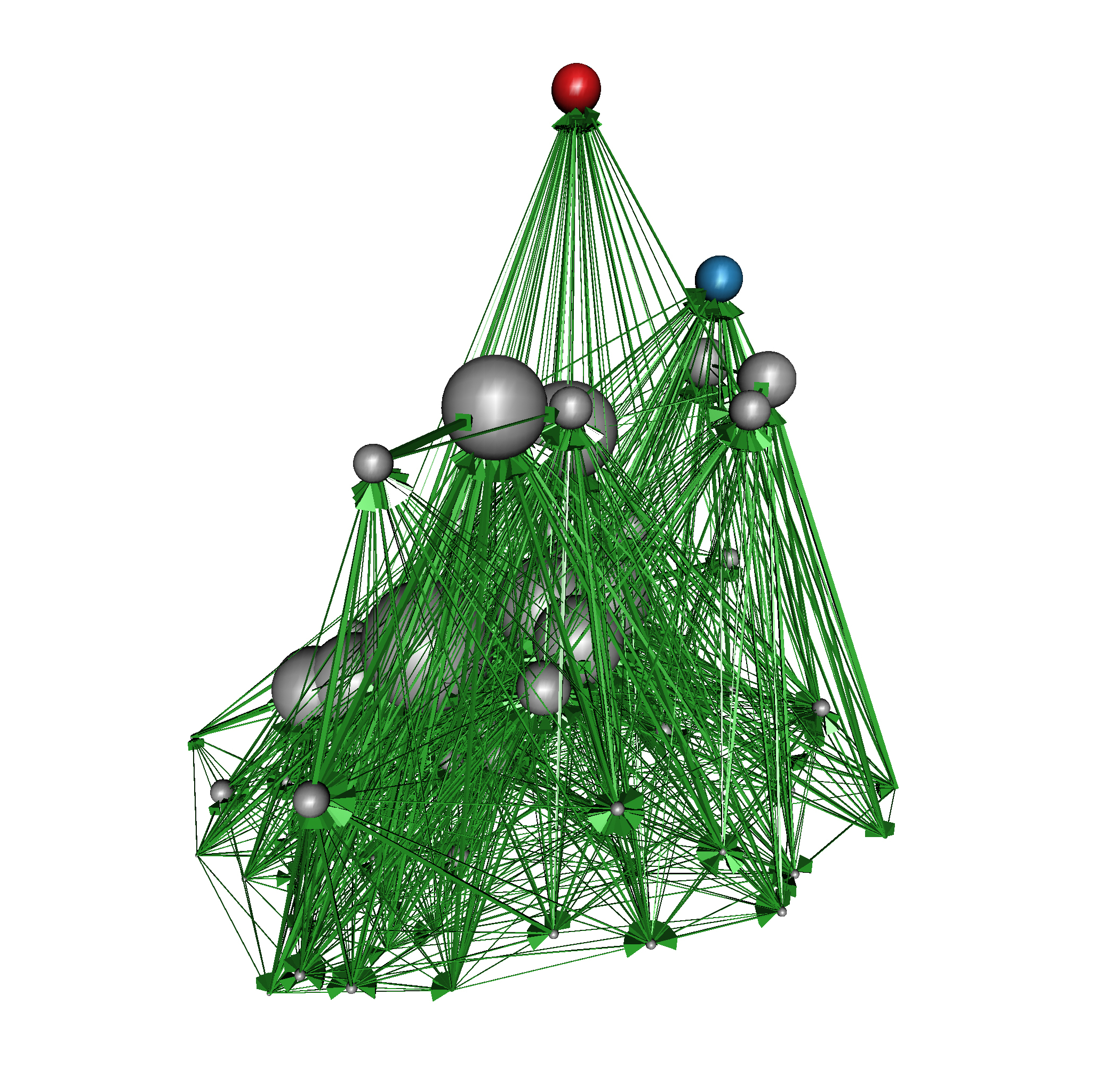}}
	\caption{3-dimensional MLON plot for \textit{cifar10} with nodes scaled by basin size and deteriorating edges removed, orientated such that the positive z-axis, corresponding to increasing fitness, points up.}
	\label{fig:cifar10graphB}
\end{center}
\vskip -0.2in
\end{figure}

Remarkably, the only data set producing a suboptimal funnel was \textit{cifar10}, depicted in \fref{fig:cifar10graphB}. The density of the graph is a result of the high modality pointed out in \sref{sec:res:nodeedge}. Despite the range of the fitness landscapes that have been populated, therefore, each contains a simple funnel structure and is therefore suitable for local search optimisation. This is a significant indication of the potential effectiveness of using LONs for architecture optimisation.

\subsubsection{Local Optimisation}

\begin{figure}[t]
\vskip 0.1in
\begin{center}
\centerline{\includegraphics[width=\columnwidth]{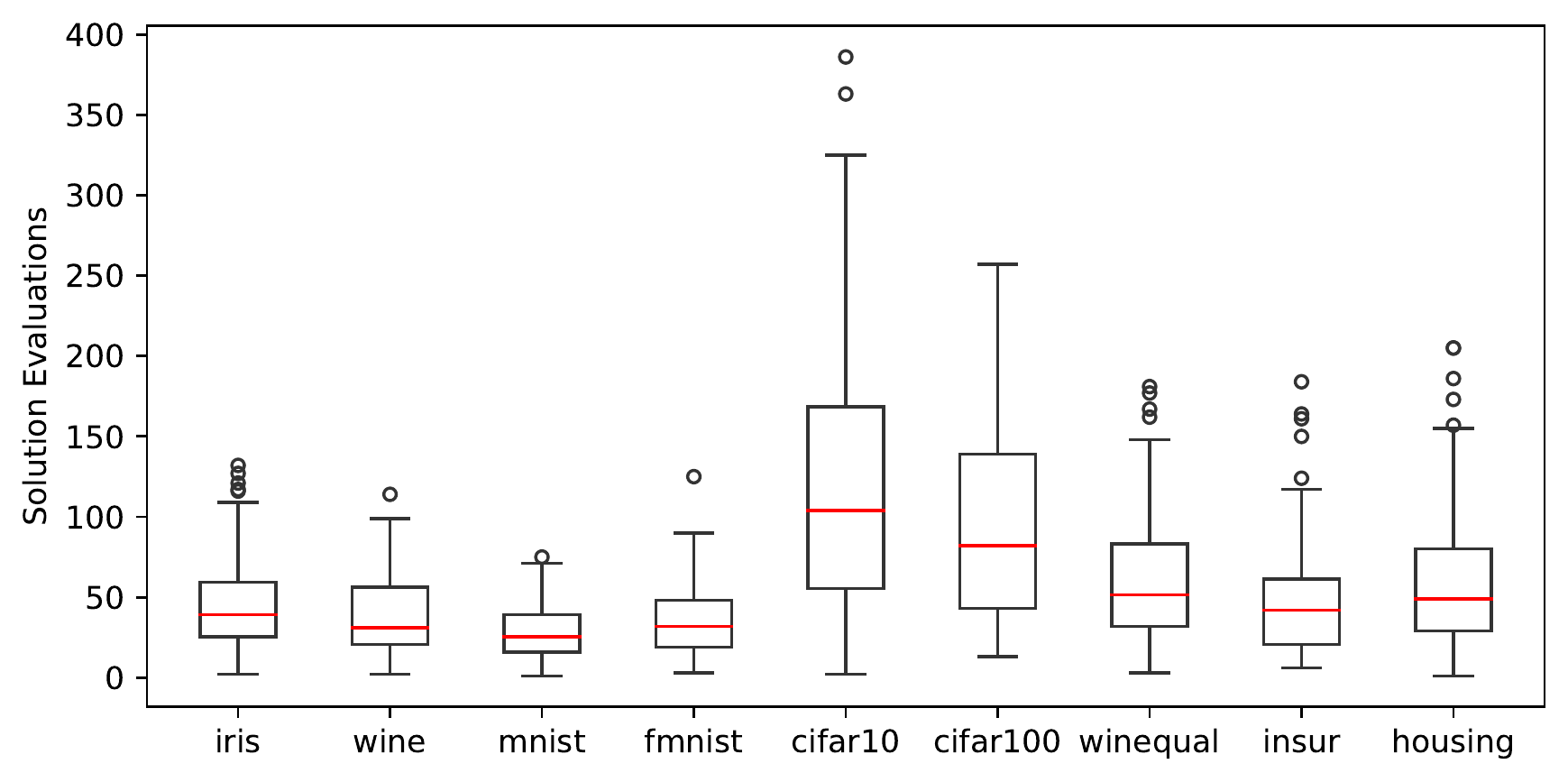}}
	\caption{First discovery of a global top 5 solution over 100 ILS procedures per data set.}
	\label{fig:prog}
\end{center}
\vskip -0.2in
\end{figure}

As an indication of fitness landscape searchability, the performance of an iterated local search (ILS) procedure, described in Algorithm~\ref{alg:ils}, was evaluated. The perturbation is defined as the same operation used to draw LON edges (\sref{sec:lons:lon}), with perturbation strength $k=2$ and stopping criterion $t=20$. Aggregating the results of 100 runs per data set, ILS discovered a global top 5 solution after evaluating on average only 50.67 evaluations (the average of the median evaluations needed per data set). \fref{fig:prog} provides a comparison of first discoveries of near optimal solutions. For almost all data sets, the median first encounter with a top 5 solution is within 50 evaluations, while the comparative hardness of \textit{cifar10} is clear. Each run only reached a fraction of all solutions before exhausting escape possibilities. Despite the limited reach, 65\% of overall runs produced the global optimum, after an average of 92.56 evaluations, which is less than a tenth of the solution space.

\begin{algorithm}
   \caption{Iterated local search (ILS)}
   \label{alg:ils}
\begin{algorithmic}
	\STATE Let $S$ be the search space; $f$ the fitness function
	\STATE Let $k$ be a perturbation strength; $t$ a stopping threshold
	\STATE Select initial random solution $s_0 \in S$
	\STATE $s \leftarrow$ \textsc{HillClimb}($s_0$)
	\STATE $i \leftarrow 0$
	\REPEAT
		\STATE $x \leftarrow$ \textsc{Perturb}($s, k$)
		\STATE $s' \leftarrow$ \textsc{HillClimb}($x$)
		\IF {$f(s') \geq f(s)$ }
			\STATE $s \leftarrow s'$
			\STATE $i \leftarrow 0$
		\ENDIF
		\STATE $i \leftarrow i + 1$
	\UNTIL {$i \geq t$}
	\vspace{0.1cm}
	\PROCEDURE {Perturb}{s, k}
		\STATE Let \textsc{Op} be a perturbation operation $\left( \mathcal{N}(s) \right)$
		\FOR {$j \leftarrow 1,\dots,k$ }
			\STATE $s' \leftarrow \textsc{Op}(s)$
			\STATE$s \leftarrow s'$
		\ENDFOR
		\STATE \textbf{return} $s$
	\ENDPROCEDURE
\end{algorithmic}
\end{algorithm}


\section{Conclusion} \label{sec:con:res}
This study presented an initial evaluation of the feasibility of using local optima network (LON) analysis to characterise and optimise over the feedforward neural network (FFNN) architecture space. The solution space was defined in such a way that full enumeration using multiple data sets was feasible. One of the main findings is the relatively simple global funnel structures of the resulting monotonic LONs (MLONs).

The LONs produced for the data sets considered in this study exhibited diverging characteristics in terms of their modality, fitness distribution, basin size distribution and global optimum incoming strength. All but one MLON, however, had a single global funnel, with only two funnels in the remaining local optima network. The simple funnel structure supports the use of relatively simple local optimisers like iterative local search as opposed to significantly more computationally expensive population-based methods.

A high standard deviation between local optimum fitness values was found to be related to low modality. Low modality in turn improves the relative ease of fitness optimisation. When encountered at low fitness levels, a very low standard deviation may indicate that the model is not sufficiently powerful for the given task, as is the case with the CIFAR-10 and CIFAR-100 data sets.\looseness=-1




One line of extension for future work is to investigate larger FFNN architecture spaces more representative of real-world tasks. Larger models are likely to improve overall fitness, which was found to correspond to simpler LONs and improve sampling performance. Another extension is to study LONs produced using different architectural arrangements, like recurrent or convolutional neural networks. Feedforward neural networks have insufficient representational capacity for many real-world problems, making basic network augmentation commonplace in modern machine learning.

This study focused exclusively on the architecture space, but weight initialisation is also known to play an important role in model performance \cite{he2015delving, zhou2019deconstructing}. To investigate the weight initialisation space, some form of discretisation is required. A promising option, considering the relatively low number of possible configurations, is to study combinations of initial weight signs, keeping the magnitude constant \cite{zhou2019deconstructing}. This line of investigation may contribute towards understanding what makes lottery tickets successful \cite{frankle2018lottery}.

\section*{Acknowledgement}
The authors would like to thank the Centre for High Performance Computing (CHPC) (\url{http://chpc.ac.za}) for the use of their cluster to obtain the data for this study.

\bibliographystyle{IEEEtran}
\bibliography{bibliography}


\end{document}